\documentclass{article}

    \PassOptionsToPackage{numbers, compress}{natbib}

    \usepackage[final]{tackling_climate_workshop_style}

\usepackage[utf8]{inputenc} 
\usepackage[T1]{fontenc}    
\usepackage{hyperref}       
\usepackage{url}            
\usepackage{booktabs}       
\usepackage{amsfonts}       
\usepackage{nicefrac}       
\usepackage{microtype}      
\usepackage{xcolor}
\usepackage{multirow}
\usepackage{subcaption}
\usepackage{graphicx} 

\title{CC-GRMAS: A Multi-Agent Graph Neural System for Spatiotemporal Landslide Risk Assessment in High Mountain Asia}

\author{%
  Mihir Panchal \\
  Department of Computer Engineering\\
  Dwarkadas Jivanlal Sanghvi College of Engineering\\
  Mumbai, India \\
  \texttt{mihirpanchal5400@gmail.com} \\
  \And
  Ying-Jung Chen \\
  College of Computing \\
  Georgia Institute of Technology \\
  Atlanta, GA 30332 \\
  \texttt{yingjungcd@gmail.com} \\
  \AND
  Surya Parkash \\
  Geo-Hydro Meteorological Risks Management Division \\
  National Institute of Disaster Management \\
  Delhi, India \\
  \texttt{surya.nidm@nic.in} \\
}

\begin{document}

\maketitle

\begin{abstract}
  Landslides are a growing climate induced hazard with severe environmental and human consequences, particularly in high mountain Asia. Despite increasing access to satellite and temporal datasets, timely detection and disaster response remain underdeveloped and fragmented. This work introduces CC-GRMAS, a framework  leveraging a series of satellite observations and environmental signals to enhance the accuracy of landslide forecasting. The system is structured around three interlinked agents Prediction, Planning, and Execution, which collaboratively enable real time situational awareness, response planning, and intervention. By incorporating local environmental factors and operationalizing multi agent coordination, this approach offers a scalable and proactive solution for climate resilient disaster preparedness across vulnerable mountainous terrains.
\end{abstract}

\section{Introduction}

Landslides, driven by climate change, pose significant risks to life and economic stability, particularly in High Mountain Asia (HMA), where complex topography, active seismicity, and shifting precipitation patterns amplify vulnerability \cite{merzdorf2020climate,dubey2023mass}. More than 1.5 billion people depend on the glaciers and rivers of the HMA, making the impacts of these hazards particularly severe \cite{baral2023climate,zhao2023new,wang2024disaster}. Key climate change drivers, such as extreme rainfall, glacial retreat, freeze thaw cycles, and permafrost degradation, exacerbate the frequency and intensity of landslides \cite{kirschbaum2020changes, shrestha2025landslides}. Furthermore, increasing population pressure, expanding infrastructure, and systemic challenges such as limited early warning systems and fragmented disaster response mechanisms increase the region's exposure to these catastrophic events \cite{manchado2022deforestation,zhou2021geoinformation, mallick2021risk}. Current approaches to landslide risk management in HMA reveal critical technical and operational gaps \cite{parkash2025assessing, kaur2024assessing, singh2019need}. Technically, spatial modeling remains weak, with poor integration of multi source data, limited incorporation of contextual knowledge, and reliance on reactive systems \cite{roccati2021gis, ma2025dynamic, liu2021landslide}. Operationally, challenges include inadequate coordination among stakeholders, low scalability of existing frameworks, insufficient real time monitoring, and weak integration of local and regional actors. These gaps hinder effective prediction, planning, and response, leaving HMA communities vulnerable to increasing climate induced landslide risks and underscoring the need for innovative, scalable solutions \cite{deng2021developing, shah2023institutional}.

To address these challenges, we propose the Climate Change Graph Risk Management and Analysis System (CC-GRMAS), a novel framework leveraging advanced AI techniques for proactive landslide risk management. CC-GRMAS employs a multi agent architecture integrating Graph Neural Networks (GNNs) with attention mechanisms for enhanced spatial modeling, Retrieval Augmented Generation (RAG) enhanced graph databases for knowledge aware decision making, and automated systems for proactive hotspot detection and real time intervention. The framework integrates data sets such as NASA's Global Landslide Catalog (1,558 HMA events) and supports Sustainable Development Goals (SDGs) 13, 11, and 15 through its modular scalable design. By combining data preprocessing, GNN based prediction, GraphRAG pipelines, and multi agent coordination, CC-GRMAS addresses systemic gaps, offering a robust solution for mitigating landslide risks in HMA.

\section{Dataset and Preprocessing}

The dataset used in this study was derived from the NASA Global Landslide Catalog (GLC), curated by the NASA Goddard Space Flight Center in collaboration with the Global Precipitation Measurement (GPM) mission. This catalog compiles landslide event records from a wide range of sources, including news articles, scientific literature, government reports, and citizen science contributions, focusing on landslides from 2007 to 2020 \cite{dandridge2023spatial}. Data were obtained through the NASA Landslide Portal (\footnote{\url{https://gpm.nasa.gov/landslides/}}, \footnote{\url{https://landslides.nasa.gov/viewer}})

\begin{figure}[h]
    \centering
    \begin{subfigure}[b]{0.5\textwidth}
        \centering
        \includegraphics[width=\textwidth]{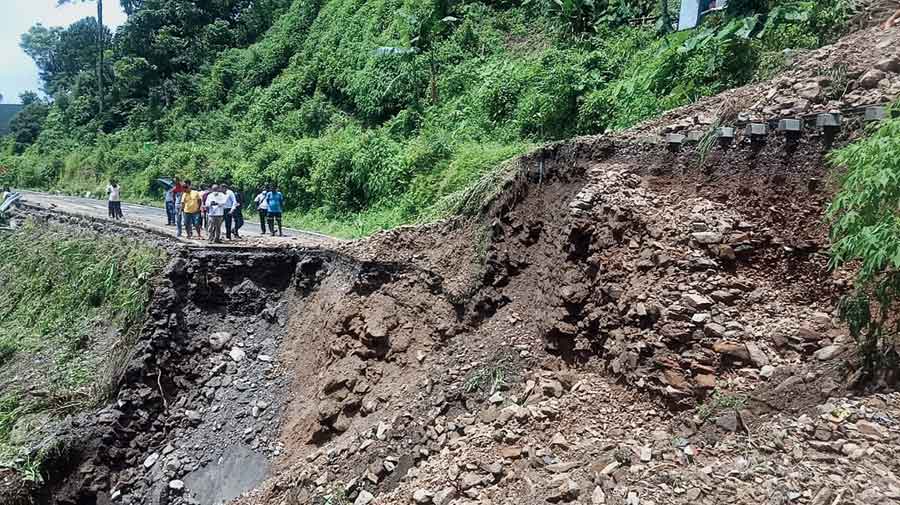}
        \caption{Landslides in High Mountain Asia}
    \end{subfigure}
    \hfill
    \begin{subfigure}[b]{0.44\textwidth}
        \centering
        \includegraphics[width=\textwidth]{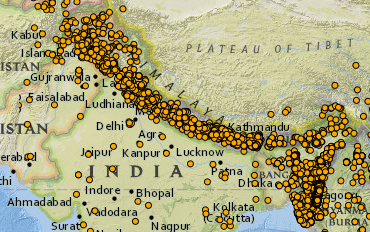}
        \caption{1558 ground truth data points}
    \end{subfigure}
\end{figure}

To ensure data consistency and usability for subsequent modeling, a series of preprocessing steps were performed. The dataset consisted of 1558 landslides events across in the HMA region. Event metadata like landslide profile, source of information and gazeteer point of the landslide were aligned with a structured knowledge graph schema to support the multiagent system. The processed dataset was then represented in a graph database, enabling seamless integration of spatial, temporal, and descriptive attributes for downstream analytical tasks.

\begin{table}[h]
  \caption{Graph Database Node Type Distribution}
  \label{tab:node_distribution}
  \centering
  \begin{tabular}{llll}
    \toprule
    Node Type & Count & Percentage & Description \\
    \midrule
    Event & 1,558 & 61.1\% & Core landslide event records \\
    Source & 440 & 17.2\% & Information sources and references \\
    GazetteerPoint & 331 & 13.0\% & Geographic reference points \\
    LandslideProfile & 223 & 8.7\% & Landslide characterization profiles \\
    \midrule
    \textbf{Total} & \textbf{2,552} & \textbf{100.0\%} & \textbf{Complete node inventory} \\
    \bottomrule
  \end{tabular}
\end{table}

The graph database organizes records into several distinct node types. Table~\ref{tab:node_distribution} summarizes the distribution of nodes in the dataset. Further details of the dataset attributes are provided in Appendix~\ref{appendix:dataset}.

\section{Proposed Method}

This study develops CC-GRMAS, a multi-agent system using graph neural networks and retrieval-augmented generation for improved landslide forecasting and disaster response in HMA. CC-GRMAS models landslide risk through graph-based spatial relationships, enhancing prediction accuracy and enabling coordinated disaster response.

The system employs an agentic architecture where each agent specializes in complementary aspects of landslide risk management. The Prediction Agent utilizes Graph Neural Networks with attention mechanisms to model spatial relationships between landslide events across geographic regions. The Planning Agent leverages Large Language Models integrated with Retrieval Augmented Generation to provide context aware risk analysis and climate impact assessments. The Execution Agent coordinates operational responses through automated hotspot detection and response generation workflows.

\begin{figure}
  \centering
  \includegraphics[width=\textwidth]{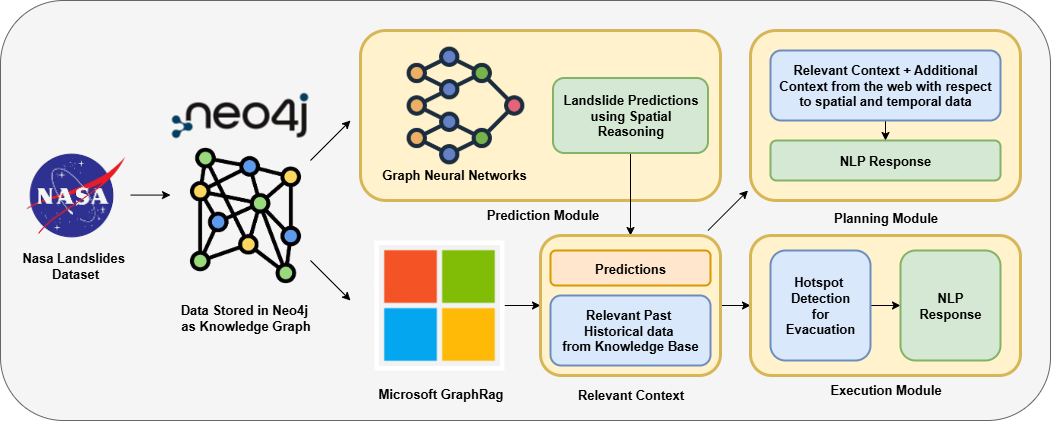}
  \caption{Architecture Diagram}
  \label{Architecture Diagram}
\end{figure}

We applied feature engineering, including spatial coordinate normalization, temporal encoding, and impact severity quantification, to improve training diversity and regional variation modeling. We then constructed dynamic proximity graphs using distance-based connectivity patterns to enhance spatial relationship modeling for landslide risk assessment. CC-GRMAS will be evaluated against NASA’s Global Landslide Catalog to validate performance across diverse geographic and temporal scenarios.\cite{kirschbaum2010global, kirschbaum2015spatial}.

The workflow of the CC-GRMAS multi agent system is shown in Figure \ref{Architecture Diagram}. This approach demonstrates that while individual machine learning models offer valuable capabilities \cite{badola2025landslide}, multi agent coordination with graph based knowledge integration is essential for optimizing performance in complex applications, such as climate induced landslide risk management \cite{kong2023landslide, zhang2024deep, ma2023knowledge}. Details of the agentic architectures and implementation  are provided in the Appendix \ref{appendix:mas}. The implementation code is available on GitHub\footnote{\url{https://github.com/MihirRajeshPanchal/CC-GRMAS}}.

\section{Conclusion \& Pathways to Climate Impact}

This work introduces CC-GRMAS, a multi-agent system for landslide forecasting and disaster response in HMA. It combines graph neural networks for spatial risk prediction, large language models using retrieval-augmented generation for contextual analysis, and automated response coordination, leveraging localized environmental data and dynamic graph representations to enhance the accuracy and timeliness of landslide risk assessments. CC-GRMAS provides real-time situational awareness, targeted planning, and rapid interventions for vulnerable communities, with a modular design adaptable to various geographies and hazards. It supports SDG 13 (Climate Action) by mitigating disaster impacts, SDG 11 (Sustainable Cities and Communities) by enhancing resilience, and SDG 15 (Life on Land) by protecting ecosystems. The framework equips policymakers and stakeholders with actionable insights to reduce risk and safeguard livelihoods in a changing climate.

\bibliography{references}
\bibliographystyle{unsrt}

\newpage
\begin{appendix}

\section{Appendix}
\subsection{Detailed Dataset Description}
\label{appendix:dataset}

The NASA GLC dataset undergoes continuous updates as new events are reported and verified, adhering to a standardized schema to ensure global comparability and high spatial precision of the data. Each landslide record encompasses a set of attributes across the temporal, spatial, impact related, and descriptive dimensions. Temporal attributes include the date of the event and the date the record was submitted, whereas spatial attributes document the latitude and longitude in decimal degrees, accompanied by a textual location description and an assessment of coordinate accuracy. Impact attributes capture the number of confirmed fatalities and the number of reported injuries. Descriptive fields provide a brief event title and a more detailed narrative of the event. The dataset's completeness and consistency render it well suited for integration into graph based representations, in which each event is linked to sources, geographic references, and landslide profiles.

\begin{table}[h]
  \caption{Landslide Event Record Attributes}
  \label{tab:event_attributes}
  \centering
  \begin{tabular}{lll}
    \toprule
    Category & Attribute & Description \\
    \midrule
    \multirow{2}{*}{Temporal} & event\_date & Date of landslide occurrence \\
     & submitted\_date & Record submission date \\
    \midrule
    \multirow{4}{*}{Spatial} & latitude & Geographic coordinate \\
     & longitude & Geographic coordinate \\
     & location\_description & Textual location description \\
     & location\_accuracy & Coordinate precision assessment \\
    \midrule
    \multirow{2}{*}{Impact} & fatality\_count & Confirmed fatalities \\
     & injury\_count & Reported injuries \\
    \midrule
    \multirow{2}{*}{Descriptive} & event\_title & Brief event identifier \\
     & event\_description & Detailed event narrative \\
    \bottomrule
  \end{tabular}
\end{table}

This structure facilitates efficient querying and enables advanced analytical techniques, such as graph neural networks. Table~\ref{tab:event_attributes} presents the complete schema for the landslide event records, including the data types and descriptions of each attribute.

\subsection{Multi Agent System Workflow}
\label{appendix:mas}

The CC-GRMAS framework addresses the fundamental challenge of landslide risk management through a coordinated multi agent architecture that integrates spatial prediction, contextual analysis, and operational response capabilities. Traditional approaches to landslide forecasting often operate in isolation, limiting their effectiveness in complex, dynamic environments where multiple factors interact across different temporal and spatial scales. The system architecture implements three specialized agents that collectively transform raw landslide data into actionable intelligence for disaster preparedness and response. Each agent addresses distinct but complementary aspects of the landslide risk management pipeline: spatial pattern recognition and prediction, contextual risk analysis and planning, and operational response coordination.

\subsubsection{Graph Neural Network Implementation Details}

\begin{figure}
    \centering
    \includegraphics[width=1\linewidth]{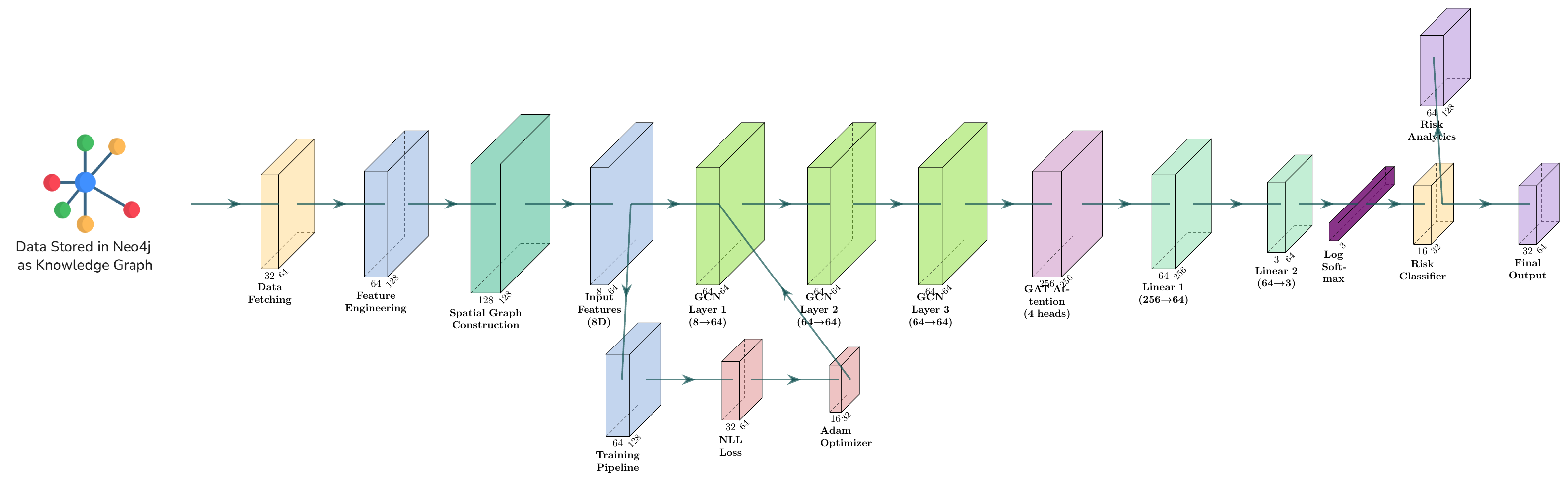}
    \caption{Detailed Graph Neural Network Architecture for the Prediction Agent showing the complete data processing pipeline from raw landslide data through feature engineering, spatial graph construction, GCN and GAT layers, to final risk classification output.}
    \label{fig:gnn_architecture}
\end{figure}

The Prediction Agent is implemented using a sophisticated Graph Neural Network architecture that processes landslide event data through multiple stages of feature engineering, graph construction, and neural network layers. Figure~\ref{fig:gnn_architecture} illustrates the complete data flow and architectural components of the GNN based prediction system.

The Prediction Agent models complex spatial relationships between landslide events across the HMA region. The agent's architecture is motivated by the observation that landslide occurrences exhibit strong spatial dependencies due to shared geological, topographical, and climatic conditions within geographic neighborhoods. The GNN implementation employs a multilayer architecture combining graph convolutional networks (GCN) with graph attention networks (GAT) to capture both local neighborhood patterns and long range spatial dependencies. The feature extraction process transforms raw landslide event data into structured node representations through spatial coordinate normalization, temporal encoding of event dates, impact severity quantification, and landslide profile categorization. These features are designed to capture the multi dimensional nature of landslide risk factors.

The spatial graph construction process creates dynamic proximity graphs using distance based connectivity patterns, where edges connect landslide events within a configurable distance threshold. This approach enables the model to learn from spatial clustering patterns while adapting to different scales of analysis. The attention mechanism within the GAT layers allows the model to automatically weight the importance of different spatial relationships, providing interpretable insights into which geographic connections most strongly influence the propagation of landslide risk. The training methodology incorporates synthetic label generation based on multifactor risk scoring that combines casualty counts, landslide magnitude classifications, and temporal seasonal patterns. This approach addresses the challenge of limited ground truth risk labels by creating training targets that reflect expert knowledge about landslide risk factors. The model outputs probabilistic risk classifications (Low, Medium, High Risk) with associated confidence scores, enabling risk aware decision making in downstream applications.

\subsubsection{GraphRAG Implementation Details}

The Planning and Execution Agents collectively implement a sophisticated knowledge integration and response generation system through GraphRAG pipelines. These agents address the critical need for contextual understanding and automated response coordination in landslide risk management scenarios. The Planning Agent leverages LLMs integrated with graph based knowledge retrieval to provide context aware risk analysis and climate impact assessments. The agent maintains a comprehensive knowledge graph representation of landslide events, their relationships, and associated metadata including landslide profiles, geographic gazetteer points, and source attribution. This structured knowledge representation enables sophisticated querying capabilities that support both semantic similarity search and graph based relationship traversal. The retrieval mechanism employs vector embeddings generated through Google Generative AI embedding models to create high dimensional representations of landslide event descriptions, location information, and contextual metadata. These embeddings are stored in Neo4j vector indexes alongside the graph structure, enabling hybrid search capabilities that combine semantic similarity with graph traversal queries. The system supports both vector similarity search for content based retrieval and Cypher query execution for structured relationship based analysis.

The generation pipeline utilizes prompt engineering techniques specifically designed for landslide risk analysis, incorporating domain specific templates that guide the language model to provide comprehensive assessments covering risk patterns, geographic distribution, temporal trends, trigger mechanisms, and climate change implications. The Execution Agent coordinates operational responses through automated hotspot detection and response generation workflows. The agent integrates predictions from the GNN-based Prediction Agent with contextual analysis from the Planning Agent to generate spatially aware risk assessments and response recommendations. The hotspot detection algorithm employs grid based spatial sampling to identify geographic regions with elevated landslide risk potential.

\subsection{Preliminary Results and Evaluation Metrics}

\subsubsection{Graph Neural Networks Implementation Evaluation}
Our Graph Neural Network (GNN) approach demonstrates significant computational advantages over traditional computer vision methodologies while maintaining slightly higher predictive performance. The spatial GNN achieves an F1-score of 0.7981 with only 42.7k parameters, representing a 99.9\% reduction in model complexity compared to the 31 million parameter 2D U-Net architecture employed in recent Nepal landslide forecasting studies~\cite{doerksen2025}. Despite this drastic reduction, our GNN matches and marginally exceeds the performance of the U-Net (0.79 F1-score), while vastly outperforming traditional machine learning baselines (0.54–0.56 F1-score).

\begin{table}[h]
\centering
\caption{Performance comparison of baseline models and U-Net for landslide forecasting.}
\label{tab:baseline_performance}
\begin{tabular}{lccccc}
\toprule
\textbf{Approach} & \textbf{Method} & \textbf{F1-Score} & \textbf{Precision} & \textbf{Recall} & \textbf{Params (M)} \\
\midrule
Nepal Study~\cite{doerksen2025} & RF   & 0.56 & 0.47 & \textbf{0.70} & $<0.1$ \\
Nepal Study~\cite{doerksen2025} & XGB  & 0.54 & 0.45 & 0.67 & $<0.1$ \\
Nepal Study~\cite{doerksen2025} & GB   & 0.56 & 0.49 & 0.65 & $<0.1$ \\
Nepal Study~\cite{doerksen2025} & U-Net & \textbf{0.79} & \textbf{0.91} & 0.69 & 31.0 \\
\textbf{CC-GRMAS (Ours)} & \textbf{Spatial GNN} & \textbf{0.7981} & \textbf{0.8062} & \textbf{0.7928} & \textbf{<0.1} \\
\bottomrule
\end{tabular}
\vspace{0.2em}
{\footnotesize *Approximate ranges estimated from performance plots in~\cite{doerksen2025}.}
\end{table}

Table~\ref{tab:baseline_performance} provides a comprehensive comparison between our spatial GNN and previously established computer vision and machine learning approaches.

This extreme reduction in parameters translates directly to faster training times, lower memory requirements, and improved deployment feasibility in resource constrained environments typical of HMA. Beyond efficiency, the graph based formulation offers greater interpretability: spatial dependencies are explicitly modeled through edge connections, while attention mechanisms highlight the critical geographic relationships driving landslide susceptibility. Importantly, scalability benefits are amplified when forecasting over larger regions, as graph representations grow linearly with the number of landslide events rather than quadratically with image resolution, making our approach more suitable for operational deployment across complex mountain terrains.

In summary, our spatial GNN model not only matches but slightly improves upon the predictive accuracy of heavy weight computer vision models while operating with an order of magnitude fewer parameters. This confirms that graph based spatial modeling provides an efficient, interpretable, and scalable solution for landslide risk assessment in HMA.

\subsubsection{GraphRAG Implementation Evaluation}

The GraphRAG pipeline integrating the Planning and Execution Agents demonstrates effective knowledge retrieval and contextual synthesis capabilities for landslide risk analysis across the High Mountain Asia region. We evaluate the system's performance through semantic coherence metrics that measure the quality of retrieved information and generated responses against ground truth data from the NASA Global Landslide Catalog.

\paragraph{Evaluation Methodology and Metrics}

The evaluation employs a semantic coherence metric that quantifies how well the retrieved nodes and generated answers align with ground truth information. The system was evaluated across multiple queries covering landslide events in Nepal, India, Pakistan, China, Bhutan, and Bangladesh, with results averaged to provide aggregate performance measures. Table~\ref{tab:semantic_coherence} presents the detailed breakdown of the semantic coherence metric and its constituent components.

\begin{table}[h]
\centering
\caption{Semantic Coherence Metric Components for GraphRAG Evaluation}
\label{tab:semantic_coherence}
\begin{tabular}{p{2.9cm} p{1cm} p{8cm}}
\toprule
\textbf{Component} & \textbf{Score} & \textbf{Description} \\
\midrule
Overall Semantic Coherence & 0.751 & Composite metric measuring retrieval quality and answer relevance averaged across all HMA countries \\
\midrule
Average Similarity & 0.814 & Mean cosine similarity between retrieved nodes and ground truth across all retrieved results \\
Weighted Similarity & 0.821 & Position weighted similarity emphasizing higher ranked retrieval results \\
Maximum Similarity & 0.840 & Best single node match indicating peak retrieval accuracy of nodes\\
Minimum Similarity & 0.797 & Weakest node match showing lower bound of retrieval quality \\
Diversity Score & 0.143 & Measure of information variety across retrieved nodes balancing relevance with coverage \\
\bottomrule
\end{tabular}
\end{table}

The semantic coherence metric is computed using vector embeddings that transform textual descriptions of landslide events into high  dimensional semantic representations. Cosine similarity measurements between these embeddings quantify the semantic alignment between retrieved nodes and ground truth references. The weighted similarity component assigns higher importance to top ranked results, reflecting user interaction patterns with retrieval systems. The diversity score captures the variance in semantic content across retrieved nodes, ensuring comprehensive coverage rather than redundant information.

The system demonstrates consistent retrieval performance across diverse geographic contexts in the HMA region. Query evaluation spanning Nepal, India, Pakistan, China, Bhutan, and Bangladesh reveals robust knowledge graph traversal capabilities and effective semantic matching across different reporting standards and source languages. The Planning Agent's answer generation pipeline successfully synthesizes information from retrieved nodes to produce comprehensive assessments covering temporal context, impact quantification, trigger identification, and geographic specificity across these varied regional contexts.

The GraphRAG approach demonstrates distinct advantages over traditional keyword based retrieval through semantic understanding capabilities and graph based relationship modeling. The vector embedding approach captures conceptual relationships beyond exact keyword matches, enabling retrieval of relevant events across different regional reporting conventions. The graph structure provides interpretable retrieval pathways that support verification and enable multi-hop reasoning to identify relationships between geographically or temporally separated events.

\paragraph{Implications and Future Directions}

The consistent performance across the six HMA countries validates the system's applicability for regional scale disaster management applications. The semantic coherence score demonstrates that the GraphRAG pipeline effectively integrates structured graph knowledge with flexible natural language generation capabilities. Future work will expand evaluation protocols to include temporal range queries, multi-region comparative analysis, and trend identification tasks to further validate operational readiness for deployment in disaster management contexts.

\end{appendix}

\end{document}